\documentclass[10pt,twocolumn,letterpaper]{article}
 \usepackage[left=2cm, right=2cm, top=2cm]{geometry} 
\usepackage{times}
\usepackage{epsfig}
\usepackage{graphicx}
\usepackage{amsmath}
\usepackage{amssymb}
\usepackage{fancyhdr}
\usepackage{amsfonts}
\usepackage{booktabs}
\usepackage{siunitx}
\usepackage[table]{xcolor}
\usepackage{bm}
\usepackage{float}
\usepackage{times}
\usepackage{subfig}

\newcommand{\pd}[2]{\frac{\partial{#1}} {\partial{#2} } }

\newcommand{\etal}{\textit{et al}.}

\fancypagestyle{plain}{%
   \fancyhf{} 
   \fancyfoot[C]{\scriptsize Authors contributed equally to this manuscript.}
   
}

\begin{document}

\title{Leveraging Class Similarity to Improve Deep Neural Network Robustness}

\author{Pooran Singh Negi \\
University of Denver\\
{\tt\small pooran.negi@gmail.com}
\and
David Chan\ \\
University of California, Berkeley\\
{\tt\small  davidchan@berkeley.edu}
\and
 Mohammad Mahoor \\
University of Denver\\
{\tt\small Mohammad.Mahoor@du.edu}
}
\date{}
\maketitle

\begin{abstract}
   Traditionally artificial neural networks (ANNs) are trained by minimizing the cross-entropy between a provided ground-truth delta distribution (encoded as one-hot vector) and the ANN's predictive softmax distribution. It seems, however, unacceptable to penalize networks equally for miss-classification between classes. Confusing the class ``Automobile" with the class ``Truck" should be penalized less than confusing the class ``Automobile" with the class ``Donkey". To avoid such representation issues and learn cleaner classification boundaries in the network, this paper presents a variation of cross-entropy loss which depends not only on the sample class but also on a data-driven prior ``class-similarity distribution" across the classes encoded in a matrix form. We explore learning the class-similarity distribution using a data-driven method and then show that by training with our modified similarity-driven loss, we obtain slightly better generalization performance over multiple architectures and datasets as well as improved performance on noisy testing scenarios.  
\end{abstract}

\section{Introduction}
\begin{figure}[h]
\centering
\includegraphics[width=0.45\textwidth,height=0.55\textwidth,clip,trim={0cm, 0cm, 0.0cm, 0cm}]{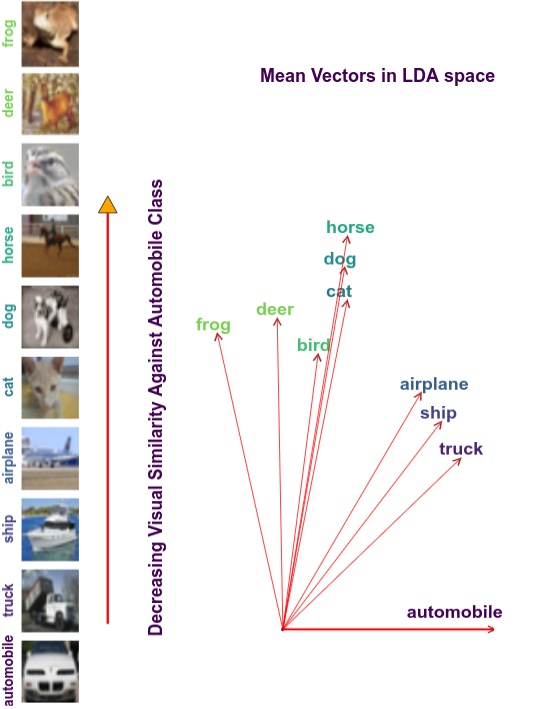}
\caption{An example of visualizing between-class similarity in a classification space. Vectors are generated (rooted at the origin) using a toy similarity measure (Section \ref{section:lda}) based on LDA.}
\label{fig:lda_cosine_sim_measure_wacv}
\vspace{10pt}
\end{figure}

	Deep neural networks are responsible for a large number of recent breakthroughs in computer vision applications - particularly in object recognition, where in some cases they even exceed human-level performance.  Traditionally, training neural networks relies on computing a loss function in the form of the cross-entropy between a predictive softmax distribution as an interpretation of the last layer of features and a  Dirac delta distribution (encoded as a one-hot vector) which encodes the ground-truth class representation of the labels ~\cite{de2005tutorial,russell2003artificial}. 
	
	Training with respect to the softmax output and ground truth delta distribution is a provably correct assumption when training a maximum likelihood classifier given that the classes observed in the training data make up the entire possible set of classes ~\cite{russell2003artificial}. While it is inadvisable to completely ignore traditional delta distribution loss in the traditional classification scheme, it is reasonable to ask if it would be possible to borrow ideas from a more general case, in which classes are considered sub-spaces of a larger image space. In this more general world, it is natural to think that two classes may be closer together in the image space. This leads to the intuition that it is inherently more likely for a classifier to ``confuse" images from similar classes and some classes may have high inter-class mutual information.
	
	For a network to properly minimize the cross entropy loss for a given class with a softmax predictive distribution requires that the output logit function for the class be infinitely large. Under the assumption of high inter-class mutual information (such as in the near-class scenario), this requirement is no longer a reasonable assumption.
	In addition, trying to approximate infinite logits leaves something to be desired. As computers are finite-representational, using back-propagation to push the class logits beyond the inherent representation capacity of the computer is computationally futile. Even further, the traditional cross-entropy loss formulation leads to large variations in weight changes during back-propagation for examples from classes that are difficult to distinguish from other classes (``near-class" examples) as well as for examples which have been ambiguously annotated. These variations lead to distortion along the classification boundary which can detrimentally affect the training procedure and the overall generalization capacity of a neural network.
	
    We might resolve these issues in one of two ways: 1) we could abandon the odds-likelihood approach to classification, which seems unnecessary (and perhaps incorrect), or 2) we can explore loss functions that do not try to approximate a ground-truth Dirac distribution. Thus we ask: ``is training the network to exactly approximate the ground truth necessarily the best option?'' In order to investigate this question we explore a new class of loss functions that are sensitive to the challenges presented above and learned in a data-driven manner. Our major contributions in this paper are:
	 
	\begin{enumerate}
      \item We introduce a ``Mixed Cross Entropy Loss'' (MCEL) framework, a new loss function which takes into account the ground truths distribution while simultaneously allowing for a certain flexibility for the network to make other predictions on the basis that such predictions may not be the fault of the network itself, but result of some similar feature or high mutual-information between different classes in the dataset. MCEL employs a convex combination of the traditional cross-entropy between the predictive softmax distribution and ground truth Dirac delta distribution and a secondary cross-entropy between the predicted distribution and ``class-similarity distribution."
      \item We explore how to learn the class-similarity distribution from the data using Linear Discriminant Analysis (LDA).
      \item We use MCEL across multiple datasets and network architectures and empirically show improved top-k error, a predictable side-effect of the MCEL formulation without significant tradeoff to top-1 error.
     \end{enumerate}
     
 	Though our exploration of the data-dependent prior belief is only a basic exposition and exploration of the MCEL framework presented in this paper, we provide a framework from which further research into data-driven loss functions for neural networks can proceed.
	
\section{Related Work}
	The recent success of deep neural networks has greatly benefited from research into network architecture \cite{szegedy2015going,Szegedy_2016_CVPR}, an exploration of deeper structure \cite{he2015deep,Szegedy_2016_CVPR}, the development of novel non-linearity functions \cite{clevert2015fast}, novel optimization techniques ~\cite{kingma2014adam,zeiler2012adadelta} and better regularization \cite{xie2016disturblabel,srivastava2014dropout}. This wide breadth of work lacks significant research on the loss function, which is usually comprised of a sole cross-entropy loss function \cite{de2005tutorial} between ground truth and the predictive softmax distribution.
	
	Manipulating the loss function in a data-driven manner has most significantly been applied to training neural networks with imbalanced data ~\cite{zhou2006training,Szegedy_2016_CVPR,xie2016disturblabel}, however in practice such techniques are not applied to improve accuracy in the general case. Indeed, using a data-driven loss function for such applications intuitively makes sense as imbalanced data requires some penalty loss function in order to compensate for the insufficient training sample from some classes.
	
	One of the most wide-reaching data-driven loss functions is Infogain loss. First implemented in Caffe ~\cite{jia2014caffe}, and thereafter adopted by the technical community, yet never published as a standalone idea, the Infogain loss formulation was developed to solve the problem of learning in imbalanced data. By individually weighting the loss function for each of the classes by their relative proportion in the training data-set, we can reduce some of the incongruities introduced by heavily penalizing networks for miss-classifying examples from under-represented classes, while penalizing networks less for miss-classifying examples from well represented classes. The loss formulation for a training example $(X,y)$ is given in Equation \eqref{eq:basic-infogain-loss} where $H$ is an Infogain matrix, $\bm{f}(x)$ the predictive softmax distribution and $k$ the number of classes.
	\small
    \begin{align}\label{eq:basic-infogain-loss}
      l(X,y) = &-\sum\limits_{i}^k  H_{i,y}\log\left({\bm{f}(X)}_i \right)
    \end{align}
    \normalsize
    
    Infogain loss relies on the ``Infogain'' matrix $H$ which traditionally consists of hyper-parameter entries in the diagonal. Additionally in some formulations diagonal is the proportion of images with respect to the smallest class, that is $H_{i,i} = (\min\limits_j N_{j})/N_{i}$ where $N_i$ is the number of images representing class $i$ and $0$ off diagonal. Notice that this is inversely proportional to the number of images, allowing us to create the highest possible loss values when a target images of the worst-represented class is mis-classified. Notice that when $H=I$, the identity, we have the traditional cross-entropy loss function. 
    
    Similar in formulation to Infogain, a number of techniques are used for correcting noise in the training set. \cite{PatriniMakingDN} explores learning a class-similarity matrix which can be used to correct for noise in the labels. While \cite{PatriniMakingDN} has similar metrics to infogain using a learned data-distribution, it is restricted to correcting noise in the data as opposed to correcting for cleaner decision boundaries. Prior works such as \cite{Menon2015LearningFC,Liu2016ClassificationWN} also explore learning from corrupted labels using a class-conditional distribution but similar to \cite{PatriniMakingDN}, they assume noise models which may not apply to high inter-class mutual information resolution. 

    Presented as a subsection in \cite{Szegedy_2016_CVPR}, Label Smoothing Regularization is a method that is used to regularize the classification layer by estimating the marginalized effect of label-dropout during training. In order to regularize the classification layer, Szegedy \etal present a formulation, where the Kronecker delta based distribution ($\delta_{y,i}$) for training sample $(x,y)$ is replaced with a mixture as given in Equation \eqref{eq:label-smoothing} where $\epsilon$ is a hyper-parameter $0 \le \epsilon \le 1$ and $u(k)$ is a fixed distribution.
    \begin{equation}\label{eq:label-smoothing}
      q(k|x) = (1 -\epsilon)\delta_{k,y} + \epsilon u(k)
    \end{equation}

	In their experiments, Szegedy \etal \, use an uniform probability distribution for the distribution $u$. Our research builds on this idea by expanding the formulation to account for arbitrary distributions in a well defined manner - while simultaneously providing additional insight into why such ``label-smoothing'' works, and applying it in a generalized scenario.
	
	Developed in \cite{yu2013kl}, KLD-Regularization introduces Kullback-Leibler divergence (KLD) as a regularization term in the traditional cross-entropy loss formulation for context-dependent deep neural network hidden Markov models. Applying the KLD regularization to the original training criterion is equivalent to changing the target probability distribution from the traditional Kronecker delta target distribution to a linear interpolation of this distribution estimated from the unadapted model and the ground truth alignment of the adaptation data. KLD differs from L2 regularized loss as it constrains not the parameters of the network, but the output probability distribution. This paper provides additional theoretic frameworks for a technique such as KLD-regularization, while generalizing the ideas behind the technique to a more global scale.
	
	Additionally, some similarity-based techniques for overcoming class imbalances have been explored. ~\cite{wang2017deep} explores learning internal class similarities from data, \cite{rossfocal} explores training under sparse sets of representative information, and \cite{zhang2017range} explores penalizing the harmonic center between classes. \cite{wen2016discriminative} uses center-loss which requires additional online computation like updating the centers, and computing pairwise distances. Unlike our proposed method, none of these techniques explore data-driven loss functions or modifying the cross-entropy approach.
	
	Experimentation with augmented loss functions are not only examined in training deep convolutional neural networks (CNNs), but also in the field of semi-supervised learning. ~\cite{grandvalet2004semi} uses an augmentation to traditional loss that measures the "missing" information in unlabeled examples. 

	Recently, examinations of cross-entropy have led to interesting improvements in deep learning performance. ~\cite{littwin2015multiverse} explores modifications to cross-entropy and demonstrates improvements in transfer learning scenarios using ensembles of networks. ~\cite{xie2016disturblabel} randomly perturbs the loss layer in order to perform regularization on the trained network. 

\section{Mixed Cross Entropy Loss}

\subsection{Notation}

    In order to facilitate an understanding of the equations involved in this paper, we establish a notation wherein the type of the quantity involved will be denoted by its representation. Scalars are represented via lower case letters  $i, j, k, \cdots; \alpha, \beta, \gamma, \cdots$, vectors are represented via lower case bold letters $\bm{a,b, \cdots, e, \epsilon}, \cdots$ and matrices are represented by upper case letters $A,B, \cdots, E, \cdots$. Bold upper case letters $\bm{A, B}, \cdots$ represent vector spaces and calligraphic letters $\mathcal{A, T, \cdots}$ are used to represent sets of objects. We consistently follow similar convention for functions, where $f$ represents scalar valued functions, bold $\bm{f}$ represents vector valued functions and bold $\bm{A_{i,.}}$ represents the i'th row of the matrix $A$. The parameters $(a, b, \cdots)$ of a function are separated using $,$ as $f(a,b,\cdots)$ and hyper-parameters $(\alpha, \beta, \cdots)$ (greek letters) are enumerated after $;$ as $f(a,b, \cdots; \alpha, \beta, \cdots)$.

\subsection{Motivation and Simple Mixed Cross-Entropy Loss}

    The loss function that we present is motivated by an application of Regularized Label Smoothing presented in ~\cite{Szegedy_2016_CVPR}, however our goal is not regularization and smoothing of the labels, but to perform dynamic penalization for the classes with respect to similarity in the training set. 
    
    As mentioned, it is unnatural to assume that if two classes have high inter-class mutual information (such as ``Horse" and ``Donkey") that the network should be aggressively penalized for confusing the two (the discussion on mutual information can be found in the supplementary document). In addition, it seems more natural to highly penalize confusion in classes with low inter-class mutual information. Thus, for a training sample $(X,y)$, we consider a ``prior'' discrete similarity distribution, described by the function $r(l=b|l=y)$ representing the probability that the label of $X$ could be $b$ under a natural distribution of images given that it has the true class $y$. That is, $r(l=b|l=y) = \mathbb{P}(\text{label is b} | \text{y is true label}) \text{ for } b \neq y$. Indeed, with the ``similarity'' distribution (defined by $r$) we wish to measure the similarity between the true-label class and the other classes in the dataset. Such a similarity distribution becomes a key hyper-parameter in our formulation.
    
    We now can define the novel formulation for ``Simple Mixed Cross-Entropy Loss'' (MCEL) to be as in Equation \eqref{eq:prop-loss} where the loss for any training example $(X,y)$ is $l(X,y)$, where $\bm{f}(X)$ is the predicted distribution, $\delta_{ij}$ is the standard Kronecker delta, $r(i|y)$ is the similarity distribution, and $\epsilon \in [0 \; .5)$ is the ``mixing weight".
    \begin{align}\label{eq:prop-loss} 
      l(X,y) = &-\sum\limits_{i}^k  [(1 - \epsilon)\delta_{yi} +\epsilon r(i|y)]\log\left({\bm{f}(X)}_i \right)
    \end{align}
    
    Because we want to retain the idea that the true labels are more important than any ``prior'' beliefs, we require that the ``mixing weight" $\epsilon$ be strictly between $0$ and $0.5$. This is, however, not a harsh requirement, and exploring the relaxation of this assumption remains as a future work. In this formulation we further restrict the similarity distribution to be symmetric across each class - that is, $r(i|y) = r(y|i)$. Because we are working on discrete classification tasks, we can provide $r$, the similarity distribution, in a matrix form. Such a formulation is presented in Equation \eqref{eq:prop-loss-mat}, where the rows of matrix $A$ represent the similarity distributions $r$ for different classes. Furthermore, we constrain $A$ to have $0s$ along the diagonal, as the true class Kronecker delta distribution accounts for the case in which the predicted class is equal to the class label.
    \begin{align}\label{eq:prop-loss-mat} 
      l(X,y) = &-\sum\limits_{i}^k  [(1 - \epsilon)\delta_{yi} +\epsilon A_{y,i}]\log\left({\bm{f}(x)}_i \right)
    \end{align}
    
    It is easy to see that such a loss function is differentiable and has an easily computed derivative with respect to the input $\bm{f}(X)$. The gradient formulas are presented for application in Equation \eqref{eq:simple-gad}. 
    \begin{align}\label{eq:simple-gad} 
     \pd{l(.)}{\bm{f}(X)_j} =
    \begin{cases}
      - \epsilon \frac{A_{y,j}} {\bm{f}(X)_j}, & \text{if} \, y \ne j \\
      - \frac{ (1-\epsilon)} {\bm{f}(X)_j}, & \text{if} \, y = j
    \end{cases}
    \end{align}
    
    It is interesting to notice that the MCEL matrices (as in the formulation of Equation \eqref{eq:prop-loss-mat}) are a subset of those suggested by the Infogain formulation. However, the MCEL matrices have two major differences which distinguish them from the Infogain matrices. The first is that the matrices have significantly more structure. While an Infogain matrix is defined as arbitrary mixture of the classes (not necessarily respecting the law of probability theory), we define the MCEL matrices with each row obeying a conditional probability distribution. By doing this, we provide a structure to the formulation and ground the choice of the matrix in an extensible theory. Second, the MCEL matrices allow for a framework for learning such a similarity distribution based on the training data (explored in Section \ref{section:lda}, which is impossible in the case of Infogain (which requires that each hyper-parameter entry in the matrix be individually investigated). 

\subsection{Semi-Generalized MCEL(SG-MCEL)}

    We continue our exploration of MCEL by presenting a natural generalization of the formulation in Equation \eqref{eq:prop-loss-mat}, where the per-class similarity distributions are mixed in different proportions. We define for each class a unique mixing weight, unlike in the simple formulation which requires $\epsilon$ be constant between all the classes. We refer to this formulation as ``semi-generalized MCEL".
    Let  $\bm{\epsilon} = 
    \begin{bmatrix} \epsilon_1, \cdots, \epsilon_i, \cdots,
      \epsilon_k 
      \end{bmatrix}$  be the vector containing probability scaling
    factors for $k$ classes. Then scaled probability vector for  $i^{th}$ class is
    $\bm{p}_i =
    {\begin{bmatrix}
      \epsilon_{i}A_{i,1},
      \epsilon_{i}A_{i,2},
      \cdots,
      (1-\epsilon_{i})A_{i,i},
      \cdots,
      \epsilon_{i}A_{i,k}
    \end{bmatrix}}^T$,  where $ A_{i,i} = 1$.
    In Equation \eqref{eq:sim-measure} we set $A_{i,i} =0$ to highlight the mixing of the class-based Kronecker delta distribution and the similarity distribution in Equation \eqref{eq:prop-loss-mat}, however this no longer needs to hold in a more general scenario. 
    
    Let the function $\bm{f}$ represent a CNN based deep neural network model for visual recognition tasks. Then for a training image $X , \bm{f}(X) \in {\mathbb{R}^+}^{k}$ represents the output probability vector based on the Softmax output layer. For a training sample set $\mathcal{T}$, the new semi-generalized MCEL loss function is:
    \begin{equation}\label{eq:per_class_prior_loss}
    l( \mathcal{T}; \epsilon_i, \cdots, \epsilon_k) = \\
    -\sum_i^n {\bm{p}_{y_i}}^T \log {\bm{f}(X_i)}.
    \end{equation}
    
    where the $\log$ function is applied component wise on the output probability vector. When $\epsilon_i = \epsilon \; \forall i \in \{1, \cdots, k \}$ then Equation \eqref{eq:per_class_prior_loss} is same as Equation ~\eqref{eq:prop-loss-mat}. 
    
    Inside the sum, the derivative of the output in this case with respect to $\bm{f}(X_i)_j$ is given in Equation (~\ref{eq:semi-general-grad}) below:
    \begin{equation}\label{eq:semi-general-grad}
     \pd{l(.)}{\bm{f}(X_i)_j} =
    \begin{cases}
      -  \epsilon_{y_{i}} \frac{A_{y_i,j}} {f(X_i)_j}, & \text{if} \, y_i \ne j \\
      -   (1-\epsilon_{y_i}) \frac{A_{j, j}} {\bm{f}(X_i)_j}, & \text{if} \, y_i = j
    \end{cases}
    \end{equation}

\subsection{Generalized MCEL(GMCEL)}
    
    It is natural to propose an even more general loss function involving minimal encoding of similarity information via a $\epsilon$-mixture matrix $E_{k, k} \in {\mathbb{R}^{+}}^{k \times k} $ (Equation (\ref{eq:epsilon-mat})). We refer to this formulation as ``generalized MCEL" (GMCEL), in which each class may have a unique hyper-parameter mixture. 
    
    \begin{equation}\label{eq:epsilon-mat}
    E_{k \times k} = 
    \begin{pmatrix}
      \epsilon_{1, 1} & \epsilon_{1, 2} & \cdots &\epsilon_{1, k} \\
      \epsilon_{2, 1} & \epsilon_{2, 2} & \cdots &\epsilon_{2, k} \\
      \vdots & \vdots & \ddots & \vdots \\
      \epsilon_{k, 1} & \epsilon_{k, 2} & \cdots &\epsilon_{k, k}
    \end{pmatrix}
    \end{equation}
    
     For the generalized MCEL mixture matrix in Equation (\ref{eq:epsilon-mat}) $ \epsilon_{i, i} > \epsilon_{i,j} + c_i $ for $i \ne j$ and $\sum_{j} \epsilon_{i,j} = 1 \; \forall i$. Scalar values $c_i > 0 $ control the class $i$ probability margin with respect to other classes. In the generalized MCEL, the loss function for a training set $\mathcal{T}$ is defined as:
    {\small
    \begin{equation}\label{eq:prior_less_loss}
    l( \mathcal{T}; E_{k,k}) = 
    -\sum_i^n 
    {\begin{bmatrix}
      \epsilon_{y_i, 1},
      \epsilon_{y_i, 2},
      \cdots,
      \epsilon_{y_i, y_i},
      \cdots,  
      \epsilon_{y_i, k} 
    \end{bmatrix}} \log{\bm{f}(X_i)}
    \end{equation}
    }
    
    Notice that if $\epsilon_{i,j} = \epsilon_i A_{i,j} $, ${i \ne j}$ and if $\epsilon_{i,i} = 1 - \epsilon_i$ if $i = j$, where each $\epsilon_i \in [0 \; .5)$, then one can choose $c_i = \frac{.5 -\epsilon_i}{2}$ making Equation \eqref{eq:prior_less_loss} equivalent to Equation \eqref{eq:per_class_prior_loss} and if $\epsilon_{i,j} = \epsilon A_{i,j}$, then Equation \eqref{eq:prior_less_loss} is same as Equation \eqref{eq:prop-loss-mat}.  
    
    In the above formulations $\epsilon, \epsilon_{k}, \epsilon_{i.j}$ are tuned as hyper-parameters using the validation set. For the ``mixture matrix", there are $O(k^2)$ hyper-parameters and tuning them can be time consuming for large $k$ values. To avoid this, we adapt the loss Equations \eqref{eq:per_class_prior_loss} and \eqref{eq:prior_less_loss} into a soft constrained-based optimization problem where the networks are responsible for learning these hyper-parameters. 
    
    \subsection {Soft Constrained Optimization of Hyper-Parameters} 
    
    Using soft constraints, Equation \eqref{eq:per_class_prior_loss} becomes:
    
    \begin{align}\label{eq:per_class_prior_loss_alt}
      l( & \mathcal{T}, \epsilon_i, \cdots, \epsilon_k; \alpha, \beta) =  l( \mathcal{T}, \epsilon_i, \cdots, \epsilon_k)
      + \nonumber \\
      & \alpha \sum_i^k (\| \bm{p}_i\|_1 -1)^2 
        + \beta \| \bm{\epsilon} -.5 \|_p^p + \gamma \| \bm{\epsilon} \|_p^p
    \end{align}
    
    where $l$ is defined to be the same as Equation \eqref{eq:per_class_prior_loss} and  $p \ge 1$ is a real number. In Equation \eqref{eq:per_class_prior_loss_alt}, the $\alpha$ penalized term ensures that $\bm{p}_i$ remains a probability vector as now $\epsilon_i$ is a learned parameter, whereas $\beta$ and $\gamma$ are the penalized terms which ensure that the components of $\bm{\epsilon}$ remain within $[0 \;.5)$. In the above formulation, the minus($-$) operator between the vector and scalar is applied in a vector component-wise.
    
    It is shown below that Equation \eqref{eq:per_class_prior_loss_alt} is differentiable with respect to its input $ \bm{f}(X_i) $ and $\bm{\epsilon} $. With respect to $\bm{f}(X_i)$, the derivative of the loss is: $$\pd{l(.)}{\bm{f}(X_i)} = -\sum_i^n \bm{p}_{y_i}^T [D(\bm{f}(X_i))]^{-1}$$ where $D(.)$ denotes a diagonal matrix with components of vector $\bm{f}(X_i)$, which lies in the diagonal of a $k \times k$ matrix. \\
    
    For $\bm{\epsilon} >0$, the derivative of the loss is:
    {\footnotesize
    \begin{align*}
    \pd{l(.)}{\bm{\epsilon}} =  - \sum_i^n [\log f(X_i)]^T D(A_{i,1}, \cdots,  -A_{y_i, y_i}, \cdots, A_{k,k}) \\
     + 2 \alpha \sum_i^k  (\|\bm{p}_i \|_1 -1)\bm{A_{i, .}} + \beta  p  \left((\bm{\epsilon} -.5) \circ (|\bm{\epsilon} -.5|)^{-1} \right)  \|\bm{\epsilon} -.5 \|^{p-1}_{p-1} \\
     + \gamma \,  p \, \left(\bm{\epsilon} \circ |\bm{\epsilon}|^{-1} \right)
      \,  \|\bm{\epsilon} \|^{p-1}_{p-1}    
      \end{align*}
      }
    where $\circ$ is the Hadamard product and $|. |^{-1}$ (absolute value and inverse) is applied component-wise on the vector.  
    Similarly, Equation \eqref{eq:prior_less_loss} becomes:
   
    \begin{align}\label{eq:prior_less_loss_alt}
      l( \mathcal{T}, E_{k,k}; & \alpha, \beta, \gamma, \eta) =
       l( \mathcal{T}, E_{k,k})
      + \alpha \sum_i^k ( \sum_j^k \epsilon_{i,j} -1)^2 \nonumber \\
      & + \beta \| E_{k,k} -1 \|_p^p + \gamma \| E_{k,k} \|_p^p \nonumber \\
      & + \eta \sum_i^k ( (k-1)(\epsilon_{i, i} -c_i)  -\sum_{j \ne i}^k\epsilon_{i,j} )^2
    \end{align}
    
    with $l$ on the right hand side same as in Equation \eqref{eq:prior_less_loss}. In Equation \eqref{eq:prior_less_loss_alt} the $\alpha$ penalized term ensures that the rows are probability vectors. For $p=2$ (Frobenius Norm case $\| . \|_F$ ),  the $\beta$ and $\gamma$ penalized terms $\| E_{k,k} -1 \|_p^p  = \sum_i^k  \sum_j^k (\epsilon_{i,j} -1)^2, \quad \| E_{k,k} \|_p^p  = \sum_i^k  \sum_j^k (\epsilon_{i,j})^2$ ensure that $\epsilon_{i,j}$ remains between $(0 \;1 )$. The last term is used to control the probability margins for different classes. 
    
    In Equations \eqref{eq:prior_less_loss_alt} and \eqref{eq:per_class_prior_loss_alt}, $\alpha, \beta, \gamma$ and $\eta$ are scalars penalizing the vector or matrix norms, hence they act as hyper-parameters controlling the space of respective parameters, {\it i.e. } acts as a regularization hyper-parameter. 
    
\section{Learning the Similarity Distribution}
\label{section:lda}

    So far, the formulation for MCEL (and even GMCEL) is left rather broad. The best way to learn a similarity distribution
    (the matrix $A$ in our formulation) for MCEL remains open.  A natural distribution can be derived from inherent class 
    variances based on linear discriminant analysis (LDA). Consider a training  set
    $\mathcal{T} = \{(X_1, y_1), \cdots, (X_m, y_m) \}$ where $X_i$ is input image(feature) and $y_i$ is the 
    corresponding class label. Let $\mathcal{T}_i = \{(X, y) | y = i \; \text{and} \; (X,y) \in \mathcal{T}\} $ be the 
    subset of training examples corresponding to class $i$. Linear discriminant analysis finds a basis set $\mathcal{B}$ for
    this labeled training set, which maximizes the inter-class variance. Not only it provides this basis set and the 
    associated projection matrix $L$, but it also gives us the components of $\mathcal{B}$ in ascending order from the 
    largest to the smallest variance (that is $\bm{b}_1$ captures more inter-class variance than $\bm{b}_2$). We project the
    training set $\mathcal{T}$ into the subspace $\bm{B}_\lambda$, spanned by the first $\lambda$ components of 
    $\mathcal{B}$. From the projected samples in each class $1 \le i \le k$ ($k$ is the number of classes), we compute the 
    average per-class vector $\bm{v_i} = \frac{1}{\#\mathcal{T}_i} \sum_{(X, y) \in \mathcal{T}_i} L X$, where 
    $\#\mathcal{T}_i$ represents the number of training samples from class $i$ and $L$ is the projection matrix. To build the similarity distributions, we 
    fill a $k \times k$ matrix $A$ with the values $A_{i,j}$ defined by:

     \begin{align}\label{eq:sim-measure} 
    A_{i,j} = \left\{ \begin{array}{cc} \frac{s(\bm{v}_i,\bm{v}_j)}{\sum_j^k s(\bm{v}_i,\bm{v}_j)} & i\neq j \\
    0 & i=j \end{array}\right.
     \end{align}
    Where: 
     \begin{align*}
     d(\bm{v}_i,\bm{v}_j) &= 1 - \frac{\langle \bm{v}_i,\bm{v}_j \rangle}{||\bm{v}_i||_2\: ||\bm{v}_j||_2}\\
     s(\bm{v}_i,\bm{v}_j) &= \frac{1}{1+e^{d(\bm{v}_i,\bm{v}_j)}} \; \text{if} \; i\ne j \nonumber
     \end{align*}
    In the above, $\langle . \, , \, . \rangle$ represents the usual vector dot product, and $s(. \, , \, .)$ defines a similarity measure between the average class vectors in the $\bm{B}_\lambda$ subspace using the cosine of the angle between the vectors as shown in Equation \eqref{eq:sim-measure}. There exists many choices for the function $s( . \, , \, .)$ as far as it is a strictly monotonically decreasing function of $d( . \, , \, . )$. Each row $\bm{A_{i}}$ of $A$  sums to one (and as $A$ is symmetric, each of the columns sum to one as well) and represents the similarity distribution for class $i$, {\it i.e.}
     $$r(\text{label} = j | \text{label} = i) = A_{i,j} \;$$
     $$ \text{with} \; \sum_j^k r(\text{label} = j|\text{label} = i) =1 \quad \forall i \in \{1, 2, \cdots, k \} $$
    Notice here that because we are using LDA, the average vector of each of the class in the subspace $\bm{B}_\lambda$ approximates the class. By taking the cosine similarity measure of these average class vectors, we build an approximate measure of how close two classes are to each other, and the closer they are to each other, the less we penalize the neural network for making a wrong prediction. Using this similarity distribution, the loss for any training example $(X,y)$, where $\bm{f}(X)$ is the predicted distribution, $\delta_{ij}$ the standard Kronecker delta and $\epsilon \in [0 \; .5)$ the mixing weight is given by MCEL directly as in Equation \eqref{eq:prop-loss-mat} (the simple MCEL formulation). We use this loss function and LDA prior for generating the empirical results in the next section.
    
    As we can see, the simple MCEL loss function replaces the true class label in the standard cross entropy based formulation by an $\epsilon$ mixture of the similarity distribution $r( . \;| \text{label} = i) = \bm{A_{i}}$ and the true class. Notice that there are some hyper-parameters in the simple MCEL formulation combined with our LDA similarity distribution. In our current experiment, $\lambda$ is equal to the number of classes $k -1$, though this is not necessary and further experimentation is required to investigate the best choices for $\lambda$. The hyper-parameter $\epsilon$ is optimized using the validation set. 
    \begin{figure}[h]
        \centering
        \includegraphics[width=0.45\textwidth,trim={0cm, 0, 2cm, 50pt},clip]{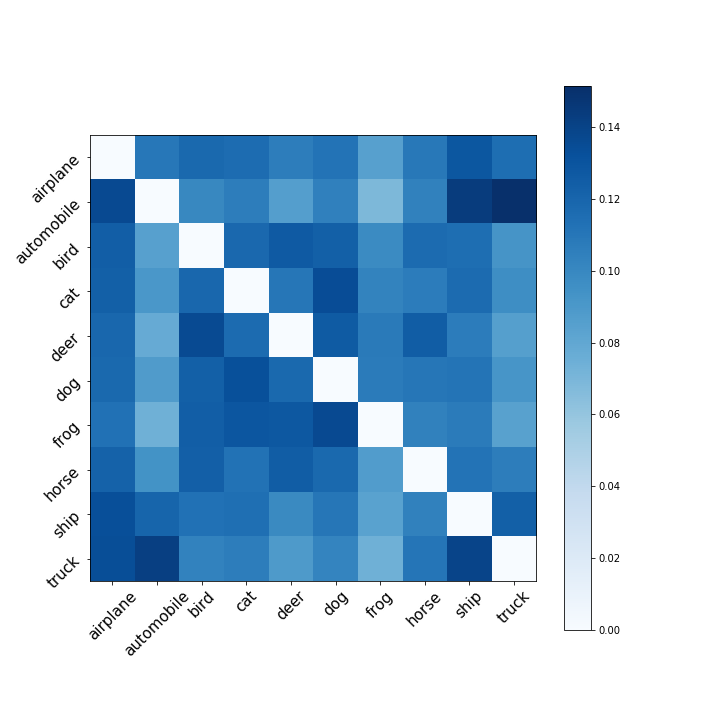}
        \caption{LDA based similarity matrix map for CIFAR-10 dataset.
         Diagonal entries corresponds to true class.}
         \label{fig:sim-map}
    \end{figure}

    A visual analysis of the LDA based similarity measure given in Figure \ref{fig:sim-map} shows that it does indeed follow our visual intuition about objects in real world. Figure \ref{fig:sim-map} shows the similarity matrix for the CIFAR-10 dataset generated by the formulation in Equation \eqref{eq:sim-measure}. The center of the matrix reveals the similarity among animal classes whereas the inanimate object classes in the corners show strong similarity. Notice that for an image of the Automobile class miss-classified as a Truck (which is arguably an automobile), our formulation reduces the exponential penalizing effect and does not greatly alter the network weights during back-propagation. 
    
\section{Experiments}

    We empirically evaluated MCEL as well as SG-MCEL with the LDA similarity metric. To train our networks, we implemented the loss function as an extension of the Tensorflow \cite{tensorflow} machine learning package using TFlearn \footnote{http://tflearn.org/}.  We run all our experiments on NVIDIA GTX 1080 Ti GPUs with 11GB memory. 
    All training data was augmented using random crops, rotations and mirrors and normalized using mean subtraction and division by the standard deviation. We used a mini-batch size of 250 images for smaller datasets, and trained using traditional SGD+Momentum. The momentum parameter was set to 0.1/0.01 and we used weight decay with value 0.1. Our learning rate decay was inverse-exponential with respect to the number of epochs. We trained each network for 500 epochs, and used the best validation epoch to perform our final inferences. To ensure re-producibility of the numbers and figures reported in the papers, all the code will be released on GitHub.
    
    In many experiments, the mixing rate $\epsilon$ was unknown. In the case of simple MCEL (denoted MCEL) we used a grid search over the range $[0,0.5]$ with interval length $0.1$, and used the best validation performance. In the case of SG-MCEL these values were learned. Results for architectures without MCEL are from the original papers and were not reproduced on our own hardware.

\subsection{Evaluation Datasets}

    \textbf{CIFAR-10 \cite{cifar} :} The CIFAR-10 dataset is a labeled subset of the 80 million tiny images dataset, $32 \times 32$ images comprising 10 classes. The dataset contains 50,000 training images and 10,000 test images. We used a randomized set of 10,000 images from the training set in order to perform validation and hyper-parameter tuning.
    
    \textbf{CIFAR-100 ~\cite{cifar} :} The CIFAR-100 dataset is a labeled subset of the 80 million tiny images dataset, $32 \times 32$ images comprising 100 classes. The dataset contains 50,000 training images and 10,000 test images. As above we used a randomized set of 10,000 images from the training set in order to perform validation and hyper-parameter tuning.
    
    \textbf{ILSVRC 2015 ~\cite{ILSVRC15} :} The ImageNet dataset contains 1000 classes, with 1.2 million images as a training set, with 50,000 validation images. The images have $224\times224\times3$ pixels. 
    
\subsection{Evaluation Architectures}

    \textbf{ResNet 32, ResNet 110, ResNet 152  ~\cite{he2015deep}:} ``Residual Networks'' use very deep architectures to set the benchmarks for state-of-the-art deep learning in a number of tasks. By learning residual functions instead of direct functions, the simplicity of training by Backpropagation is greatly increased, and the network can be extended deeply. 
    
    \noindent \textbf{Wide ResNet ~\cite{zagoruyko2016wide}:} Wide Residual Networks are residual networks which trade the depth of the traditional ResNet for an increase the width of the residual modules. 
   
\section{Results \& Analysis}
\begin{figure}[h!]
        \centering
        \includegraphics[width=0.35\textwidth, height=0.22\textwidth,clip]{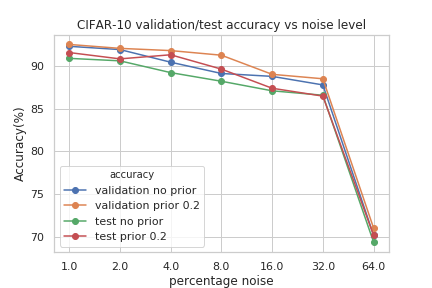}
        \caption{Training Accuracy  curves under ResNet 32 with increasing noise scenario. Percentage noise is the percentage of labels exchanged betweeen any two classes }
         \label{fig:graphs_noise_exp}
         \vspace*{-.1in}
    \end{figure}

Our experiments on the presented datasets strongly support the use of a similarity distribution prior belief in training deep neural networks. Table \ref{table:comp_param} and \ref{table:baselines} show that under our similarity distribution, we achieve promissing results on the CIFAR-10 and CIFAR-100 datasets. Because the CIFAR is a smaller dataset with easily separable classes, we expect only small gains in performance, as the generalization boundary under MCEL does not have to account for many near-class samples. To test the idea that the generalization boundaries are better laid out under simple datasets, we perform an experiment where we artificially introduce label noise into the CIFAR-10 training data. In this experiment, we randomly paired 10 classes into 5 group and swapped labels before training. The test and validation set remains true to ground truth. Figure \ref{fig:graphs_noise_exp} confirms our beliefs that under noisy data, MCEL is able to generalize better to the true distribution.
This empirically confirms that MCEL learns a more generalizable decision boundary by not strongly penalizing wrongly labelled examples.
These experiments also confirm the results of \cite{PatriniMakingDN}, which performs similar experiments, though instead of learning the similarity matrix from data, they build it directly from a noise model. 
\begin{table}[h]
\centering
\resizebox{.5\linewidth}{!}{
    \begin{tabular}{lrr}\toprule
        Method & CIFAR-10 & CIFAR-100  \\ \midrule
        WRN 28-12 & 88.28 &  59.14 \\ 
        ResNet & 90.29 & 68.24 \\ 
        WRN 28-12 + MCEL & 86.81 & 63.79 \\ 
        ResNet + MCEL & 90.44 & 67.99 \\ \bottomrule
    \end{tabular}
    }
    \vspace*{1mm}
    \caption{\small{Comparison of Top-1 accuracy (\%) on CIFAR-10 (ResNet 34) and CIFAR-100 (ResNet 152) test sets using some current methods. These use $\epsilon=0.35$.}}  \label{table:comp_param}
\end{table}


To show the generalization to more complex architectures, we evaluated on the ILSVRC-2015 dataset. Our results (Table \ref{table:imagenet}) are comparable to the state-of-the-art benchmarks in the top-k errors. 

\begin{table}[h]
\centering
\resizebox{.5\linewidth}{!}{
    \begin{tabular}{lrr}\toprule
        Method $\:$ ILSVRC 2015 &(Top-1) & (Top-5)  \\ \midrule
        ResNet 152 & 77.00 & 93.3 \\
        ResNet 152* &78.066  & 93.89\\
        ResNet 152 + MCEL & \textbf{$77.998$} & \textbf{$93.982$}\\
        ReseNet 152+SG-MCEL & 78.15 & 94.13   \\ \bottomrule
    \end{tabular}
    }
    \vspace*{1mm}
    \caption{\small{Comparison of accuracy (\%) on ILSVRC 2015 (ImageNet) validation set against easily bench-marked state-of-the-art methods. Network denoted with a '*' is the network which we trained ourselves with 0 prior weight.  }}  \label{table:imagenet}
\end{table}

We additionally show that the MCEL's performance gains are not limited to high-performance networks. Table \ref{table:baselines} shows that even under ResNet 32, a low parameter architecture MCEL can gain over a baseline architecture. It is worth noting that computing the similarity distribution takes time, however, that only needs to be computed once for each dataset and can be stored in a relatively small amount of memory (roughly 40KB to store for the CIFAR-100 dataset).
\begin{table}[h]
\centering
\resizebox{.55\linewidth}{!}{
    \begin{tabular}{lrr}\toprule
        Method & CIFAR-10 & CIFAR-100  \\ \midrule
        ResNet 32 Baseline & $92.49$ &  \\
        ResNet 32 Baseline* & $91.77 \pm 0.15$ & $63.34 \pm 0.16$ \\
        ResNet 32 + MCEL & $92.02 \pm 0.20$  & $63.72 \pm 0.10$ \\
        ResNet 32 + SG-MCEL & $91.91\pm 0.27$ &  $63.86 \pm 0.49$ \\ \bottomrule
    \end{tabular}
    }
    \vspace*{1mm}
    \caption{\small{Comparison of Top-1 accuracy (\%) on CIFAR-10 and CIFAR-100 test sets. Network denoted with a '*' is the network which we trained ourselves with $0$ prior weight. MCEL based network is optimized over prior weight($\epsilon = .2$). }}
    \label{table:baselines}
\end{table}




\begin{figure}[h!]
        \centering
        \includegraphics[width=0.35\textwidth, height=0.25\textwidth,clip]{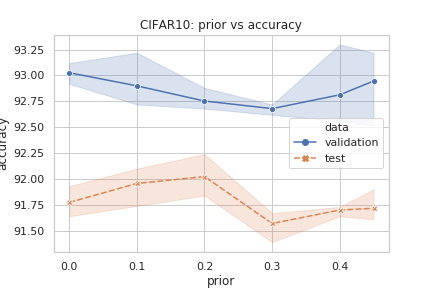}
        \includegraphics[width=0.35\textwidth, height=0.25\textwidth,clip]{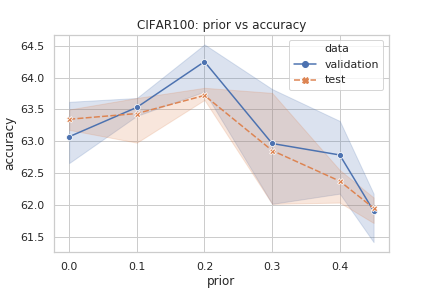}
        \caption{Accuracy vs prior  curves under ResNet 32.}
         \label{fig:acc_vs_prior}
         \vspace*{-.1in}
    \end{figure}

\begin{figure}[h!]
        \centering
        \includegraphics[width=0.35\textwidth, height=0.2\textwidth,clip]{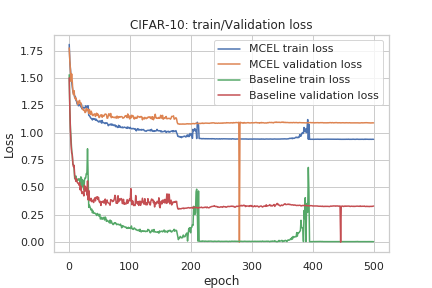}
        \includegraphics[width=0.35\textwidth, height=0.2\textwidth,clip]{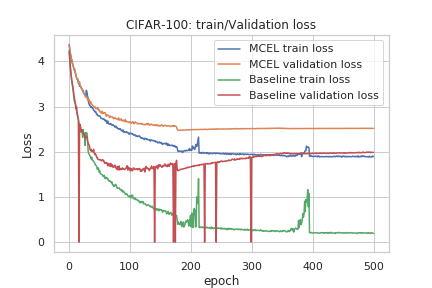}
        \caption{Training loss  curves under ResNet 32.}
         \label{fig:graphs_loss}
         \vspace*{-.1in}
    \end{figure}

Figure \ref{fig:acc_vs_prior} shows how the accuracy is affected by the choice of prior weight $\epsilon$ for various datasets in our experiments.
Figure \ref{fig:graphs_loss} shows that the convergence of MCEL networks is bit smoother than traditional cross-entropy model. Sudden drop to 0 in Loss value looks like tflearn library issue.
    
    \vspace*{-.1in}
\section{Conclusion \& Future Work}
\vspace*{-.1in}
In this work we presented a new approach, MCEL, which provides a natural extension to traditional cross-entropy based loss functions. By using MCEL as an objective function, we showed that by providing a structural guidance on the latent space (though the allowances we give to similar-class examples) leads to generalizable classification boundaries, improving generalization performance and
providing the expected results on noisy datasets. The MCEL formulation also defines a structured subset of Infogain matrices with a clear probabilistic foundation. 

While the results shown using MCEL are interesting, they are only a preliminary investigation into the power of the method for improving the robustness of deep neural networks. The similarity distribution could be learned from data in a number of different ways beyond LDA, such as by using reinforcement learning methods, multi-task learning or Siamese networks. It is additional future work to investigate what kinds of similarity measures can be learned on nonlinear manifolds beyond a pair-wise distance. 

\bibliographystyle{ieee}
\bibliography{egbib}

\begin{thebibliography}{10}\itemsep=-1pt

\bibitem{tensorflow}
M.~Abadi, A.~Agarwal, P.~Barham, E.~Brevdo, Z.~Chen, C.~Citro, G.~S. Corrado,
  A.~Davis, J.~Dean, M.~Devin, et~al.
\newblock Tensorflow: Large-scale machine learning on heterogeneous systems,
  2015.
\newblock {\em Software available from tensorflow. org}, 1, 2015.

\bibitem{clevert2015fast}
D.-A. Clevert, T.~Unterthiner, and S.~Hochreiter.
\newblock Fast and accurate deep network learning by exponential linear units
  (elus).
\newblock {\em arXiv preprint arXiv:1511.07289}, 2015.

\bibitem{de2005tutorial}
P.-T. De~Boer, D.~P. Kroese, S.~Mannor, and R.~Y. Rubinstein.
\newblock A tutorial on the cross-entropy method.
\newblock {\em Annals of operations research}, 134(1):19--67, 2005.

\bibitem{grandvalet2004semi}
Y.~Grandvalet and Y.~Bengio.
\newblock Semi-supervised learning by entropy minimization.
\newblock In {\em Advances in neural information processing systems}, pages
  529--536, 2004.

\bibitem{he2015deep}
K.~He, X.~Zhang, S.~Ren, and J.~Sun.
\newblock Deep residual learning for image recognition.
\newblock {\em arXiv preprint arXiv:1512.03385}, 2015.

\bibitem{jia2014caffe}
Y.~Jia, E.~Shelhamer, J.~Donahue, S.~Karayev, J.~Long, R.~Girshick,
  S.~Guadarrama, and T.~Darrell.
\newblock Caffe: Convolutional architecture for fast feature embedding.
\newblock In {\em Proceedings of the 22nd ACM international conference on
  Multimedia}, pages 675--678. ACM, 2014.

\bibitem{kingma2014adam}
D.~Kingma and J.~Ba.
\newblock Adam: A method for stochastic optimization.
\newblock {\em arXiv preprint arXiv:1412.6980}, 2014.

\bibitem{cifar}
A.~Krizhevsky and G.~Hinton.
\newblock Learning multiple layers of features from tiny images.
\newblock 2009.

\bibitem{littwin2015multiverse}
E.~Littwin and L.~Wolf.
\newblock The multiverse loss for robust transfer learning.
\newblock {\em arXiv preprint arXiv:1511.09033}, 2015.

\bibitem{Liu2016ClassificationWN}
T.~Liu and D.~Tao.
\newblock Classification with noisy labels by importance reweighting.
\newblock {\em IEEE Transactions on Pattern Analysis and Machine Intelligence},
  38:447--461, 2016.

\bibitem{Menon2015LearningFC}
A.~K. Menon, B.~van Rooyen, C.~S. Ong, and B.~Williamson.
\newblock Learning from corrupted binary labels via class-probability
  estimation.
\newblock In {\em ICML}, 2015.

\bibitem{PatriniMakingDN}
G.~Patrini, A.~Rozza, A.~K. Menon, R.~Nock, and L.~Qu.
\newblock Making deep neural networks robust to label noise: A loss correction
  approach.
\newblock CVPR, 2017.

\bibitem{rossfocal}
T.-Y. L. P.~G. Ross and G.~K. H.~P. Doll{\'a}r.
\newblock Focal loss for dense object detection.
\newblock {\em Proceedings of the IEEE Conference on Computer Vision and
  Pattern Recognition}, 2017.

\bibitem{ILSVRC15}
O.~Russakovsky, J.~Deng, H.~Su, J.~Krause, S.~Satheesh, S.~Ma, Z.~Huang,
  A.~Karpathy, A.~Khosla, M.~Bernstein, A.~C. Berg, and L.~Fei-Fei.
\newblock {ImageNet Large Scale Visual Recognition Challenge}.
\newblock {\em International Journal of Computer Vision (IJCV)},
  115(3):211--252, 2015.

\bibitem{russell2003artificial}
S.~J. Russell, P.~Norvig, J.~F. Canny, J.~M. Malik, and D.~D. Edwards.
\newblock {\em Artificial intelligence: a modern approach}, volume~2.
\newblock Prentice hall Upper Saddle River, 2003.

\bibitem{srivastava2014dropout}
N.~Srivastava, G.~E. Hinton, A.~Krizhevsky, I.~Sutskever, and R.~Salakhutdinov.
\newblock Dropout: a simple way to prevent neural networks from overfitting.
\newblock {\em Journal of Machine Learning Research}, 15(1):1929--1958, 2014.

\bibitem{szegedy2015going}
C.~Szegedy, W.~Liu, Y.~Jia, P.~Sermanet, S.~Reed, D.~Anguelov, D.~Erhan,
  V.~Vanhoucke, and A.~Rabinovich.
\newblock Going deeper with convolutions.
\newblock In {\em Proceedings of the IEEE Conference on Computer Vision and
  Pattern Recognition}, pages 1--9, 2015.

\bibitem{Szegedy_2016_CVPR}
C.~Szegedy, V.~Vanhoucke, S.~Ioffe, J.~Shlens, and Z.~Wojna.
\newblock Rethinking the inception architecture for computer vision.
\newblock In {\em Proceedings of the IEEE Conference on Computer Vision and
  Pattern Recognition}, June 2016.

\bibitem{wang2017deep}
J.~Wang, F.~Zhou, S.~Wen, X.~Liu, and Y.~Lin.
\newblock Deep metric learning with angular loss.
\newblock {\em Proceedings of the IEEE Conference on Computer Vision and
  Pattern Recognition}, 2017.

\bibitem{wen2016discriminative}
Y.~Wen, K.~Zhang, Z.~Li, and Y.~Qiao.
\newblock A discriminative feature learning approach for deep face recognition.
\newblock In {\em European Conference on Computer Vision}, pages 499--515.
  Springer, 2016.

\bibitem{xie2016disturblabel}
L.~Xie, J.~Wang, Z.~Wei, M.~Wang, and Q.~Tian.
\newblock Disturblabel: Regularizing cnn on the loss layer.
\newblock {\em arXiv preprint arXiv:1605.00055}, 2016.

\bibitem{yu2013kl}
D.~Yu, K.~Yao, H.~Su, G.~Li, and F.~Seide.
\newblock Kl-divergence regularized deep neural network adaptation for improved
  large vocabulary speech recognition.
\newblock In {\em Acoustics, Speech and Signal Processing (ICASSP), 2013 IEEE
  International Conference on}, pages 7893--7897. IEEE, 2013.

\bibitem{zagoruyko2016wide}
S.~Zagoruyko and N.~Komodakis.
\newblock Wide residual networks.
\newblock {\em arXiv preprint arXiv:1605.07146}, 2016.

\bibitem{zeiler2012adadelta}
M.~D. Zeiler.
\newblock Adadelta: an adaptive learning rate method.
\newblock {\em arXiv preprint arXiv:1212.5701}, 2012.

\bibitem{zhang2017range}
X.~Zhang, Z.~Fang, Y.~Wen, Z.~Li, and Y.~Qiao.
\newblock Range loss for deep face recognition with long-tailed training data.
\newblock In {\em Proceedings of the IEEE Conference on Computer Vision and
  Pattern Recognition}, pages 5409--5418, 2017.

\bibitem{zhou2006training}
Z.-H. Zhou and X.-Y. Liu.
\newblock Training cost-sensitive neural networks with methods addressing the
  class imbalance problem.
\newblock {\em IEEE Transactions on Knowledge and Data Engineering},
  18(1):63--77, 2006.

\end{thebibliography}

\end{document}